\documentclass[review]{elsarticle}

\usepackage{lineno,hyperref}
\modulolinenumbers[5]

\journal{Journal of Neurocomputing}
\usepackage{numcompress}\biboptions{authoryear}

\usepackage{amsmath}
\usepackage{amsthm}
\usepackage{amsfonts}
\usepackage{times}

\usepackage{setspace}
\usepackage[dvipsnames]{xcolor}

\newcommand{\R}{\mathbb{R}}

\newcommand{\x}{\mathbf{x}}

\newcommand{\btheta}{\boldsymbol{\theta}}

\usepackage{times}
\usepackage{soul}
\usepackage{url}
\usepackage{algorithm}
\usepackage{algorithmic}
\usepackage{amsfonts}
\usepackage{algorithm}
\usepackage{algorithmic}
\usepackage{graphicx}
\usepackage{lscape}
\usepackage{booktabs}
\usepackage{caption}
\usepackage{subcaption}
\usepackage{multirow}
\begin{document}

\begin{frontmatter}

\title{Domain Generalization via Optimal Transport with Metric Similarity Learning}

\author[mymainaddress]{Fan Zhou\corref{mycorrespondingauthor}}
\cortext[mycorrespondingauthor]{For any concerns, please contact:}
\ead{fan.zhou.1@ulaval.ca}
\author[mysecondaryaddress]{Zhuqing Jiang}
\author[mythirdaddress]{Changjian Shui}
\author[myfourthaddress,myfifthaddress]{Boyu Wang}
\author[mymainaddress]{Brahim Chaib-draa}

\address[mymainaddress]{Department of Computer Science and Software Engineering, Laval University, QC, Canada}
\address[mysecondaryaddress]{School of Information and Communication Engineering, Beijing University of Posts and Telecommunications, Beijing, China}
\address[mythirdaddress]{Department of Electrical and Computer Engineering, Laval University, QC, Canada}
\address[myfourthaddress]{Department of Computer Science, University of Western Ontario, ON, Canada}
\address[myfifthaddress]{Vector Insitute, ON, Canada}

\begin{abstract}
\noindent 
Generalizing knowledge to unseen domains, where data and labels are unavailable, is crucial for machine learning. We tackle in this chapter the domain generalization problem to learn from multiple source domains and generalize to a target domain with unknown statistics. The crucial idea is to extract the underlying invariant features across all the domains. Previous domain generalization approaches mainly focused on learning invariant features and stacking the learned features from each source domain to generalize to a new target domain while ignoring the label information, this generally leads to indistinguishable features with an ambiguous classification boundary. One possible solution is to constrain the label-similarity when extracting the invariant features and take advantage of the label similarities for class-specific cohesion and separation of features across domains. 
We adopt here the optimal transport with Wasserstein distance, which could constrain the class label similarity, for adversarial training. We also deploy a metric learning objective to leverage the label information for achieving distinguishable classification boundary. Our empirical results show that our proposed method could outperform most of the baselines. Furthermore, ablation studies also demonstrate the effectiveness of each component of our method.
\end{abstract}

\begin{keyword}
Domain Generalization \sep Adversarial Learning \sep  Metric Learning
\end{keyword}

\end{frontmatter}

%\linenumbers

\section{Introduction}
Recent years witness a rapid development of machine learning and its succeeded applications such as computer vision~\citep{ma2018variational,zhu2019image,ma2019shoe}, natural language processing~\citep{ma2013vector,ma2018short} and cross-modalities learning~\citep{zhu2019image,xu2018cross} with many real-world applications~\citep{xie2018mobile,ma2019fine}. Traditional machine learning methods are typically based on the assumption that training and testing datasets are from the same distribution. However, in many real-world applications, this assumption may not hold, and the performance could degrade rapidly if the trained models are deployed to domains different from the training dataset~\citep{ganin2016domain}. More severely, to train a high-performance vision system requires a large amount of labelled data, and getting such labels may be expensive. 
 Taking a pre-trained robotic vision system as an example, during each deployment task, the robot itself (\emph{e.g.} position and angle), the environment (\emph{e.g.} weather and illumination) and the camera (\emph{e.g.} resolution) may result in different image styles. The cost to annotate enough data for each deployment task could be very expensive.

This kind of problem has been widely addressed by transfer learning (TL)~\citep{zhuang2019comprehensive} and domain adaptation (DA)~\citep{ganin2016domain}. In DA, a learner usually has access to the labelled source data and unlabelled target data, and it is typically trained to align the feature distribution between the source and target domain. However, sometimes, we could not expect the target data is accessible for the learner.
In the robot example, the distribution divergences (different image styles) from training to testing domain can only be identified after the model is trained and deployed. 
In this scenario, it's unrealistic to collect samples before deployment. This would require a robot to have abilities to handle domain divergences even though the target data is absent. 

\begin{figure}
    \centering
    \includegraphics[width=0.99\textwidth]{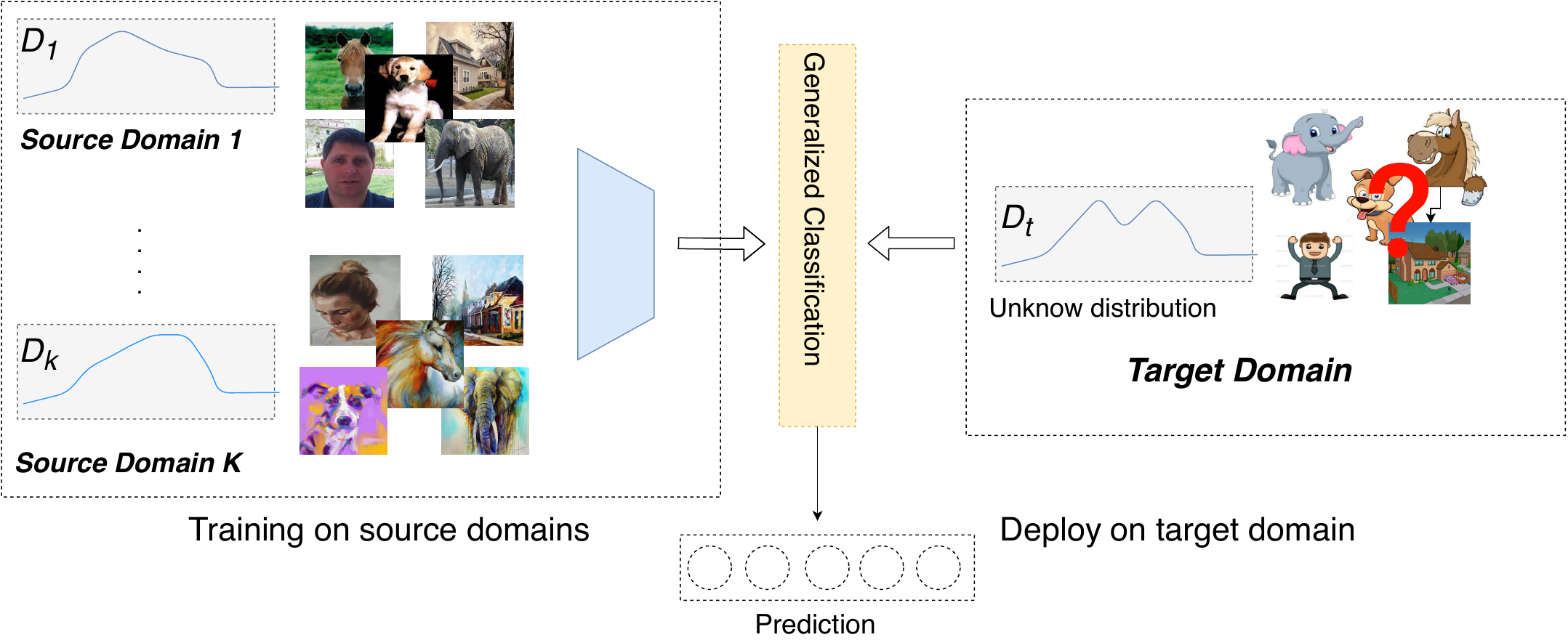}
    \caption{Domain Generalization: A learner faces a set labelled data from several source domains, and it aims at extracting invariant features across the seen source domains and learn to generalize to an unseen domain. Based on the manifold assumption~\citep{goldberg2009multi}, each domain $i$ is supported by distribution $\mathcal{D}_i$. The learner can measure the source domain distribution via the source datasets but has no information on the unseen target distribution. After training on the source domains, the model is then deployed to a new domain $\mathcal{D}_t$ for prediction.}
    \label{fig:DG}
\end{figure}

We tackle this kind of problem under domain generalization (DG) paradigm, under which the learner has access to many source domains (data and corresponding labels), and aims at generalizing to the new (target) domain, where both data and labels are unknown. 
The goal of DG is to learn a prediction model on training data from the seen source domains so that it can generalize well on the unseen target domain. 
An underlying assumption behind domain generalization is that there exists a common feature space underlying the multiple known source domains and unseen target domain. Specifically, we want to learn domain invariant features across these source domains, and then generalize to a new domain.
An example of how domain generalization is processed is illustrated in Fig.\ref{fig:DG}.

 A critical problem in DG and DA involves aligning the domain distributions, which typically are achieved by extracting such representations. 
 Previous DA works usually tried to minimize the domain discrepancies, such as KL-divergence and Maximum Mean Discrepancy (MMD) etc. via adversarial training, to achieve domain distribution alignments. Due to the similar problem setting between DA and DG, many previous approaches directly adopt the same adversarial training technique for DG. For example, a MMD metric is adopted by~\cite{li2018domain} as a cross-domain regularizer and KL divergence is adopted to measure the domain shift by~\cite{Li2017dg} for domain generalization problem.  
 The MMD metric is usually implemented in kernel space, which is not sufficient for large-scaled applications, and KL divergence is unbounded, which is also insufficient for a successful measuring domain shift~\citep{zhao2019learning}. 
 
 Besides, previous domain generalization approaches~\citep{ilse2019diva,ghifary2015domain,li2018deep,d2018domain,volpi2018generalizing} mainly focused on applying similar DA technique to extract the invariant features and how to stack the learned features from each domain for generalizing to a new domain. These methods usually ignore the label information and will sometimes make the features became indistinguishable with ambiguous classification boundaries, $a.k.a$ semantic misalignment problem~\citep{8964455}.
  A successful generalization should guide the learner not only to align the feature distributions between each domain but also to discriminate the samples in the same class could lie close to each other while samples from different classes could stay apart from each other, $a.k.a.$ feature compactness~\citep{kamnitsas2018semi}. 
 
Aiming to solve this, we adopt Optimal Transport (OT) with Wasserstein distance to align the feature distribution for domain generalization since it could constrain labelled source samples of the same class to remain close during the transportation process~\citep{courty2016optimal}. Moreover, some information theoretical metrics such as KL divergence is not capable to measure the inherent geometric relations among the different domains~\citep{arjovsky2017wasserstein}. In contrast, OT can exactly measure their corresponding geometry properties. Besides, compared with~\citep{ben-david2010}, OT benefits from the advantages of Wasserstein distance by its gradient property~\citep{arjovsky2017wasserstein} and the promising generalization bound~\citep{redko2017theoretical}. The empirical studies~\citep{gulrajani2017improved,shen2017wasserstein} also demonstrated the effectiveness of OT for extracting the invariant features to align the marginal distributions of different domains. 

Furthermore, although the optimal transport process could constrain the labelled samples of the same class to stay close to each other, our preliminary results showed that just implementing optimal transport for domain generalization is not sufficient for a cohesion and separable classification boundary. The model could still suffer from indistinguishable features (see Fig.~\ref{fig:sub-third}). In order to train the model to predict well on all the domains, this separable classification boundary should also be achieved under a domain-agnostic manner. That is, for a pair of instances, no matter which domain they come from, they should stay close to each other if they are in the same class and vice-versa. To this end, we further promote metric learning as an auxiliary objective for leveraging the source domain label information for a domain-independent distinguishable classification boundary. 

To summarize, we deployed the optimal transport technique with Wasserstein distance for domain generalization for extracting the domain invariant features. To avoid ambiguous classification boundary, we proposed to implement metric learning strategies to achieve a distinguishable feature space. Therefore, we proposed the Wasserstein Adversarial Domain Generalization (\emph{WADG}) algorithm. 

In order to check the effectiveness of the proposed approach, we tested the algorithm on two benchmarks comparing with some recent domain generalization baselines. 
The experiment results showed that our proposed algorithm could outperform most of the baselines, which confirms the effectiveness of our proposed algorithm. Furthermore, the ablation studies also demonstrated the contributions of our algorithm.

\section{Related Works}

\subsection{Domain Generalization}\noindent

 The goal of DG is to learn a model that can extract common knowledge that is shared across source domains and generalize well on the target domain. Compare with DA, the main challenge of DG is that the target domain data is not available during the learning process.

A common framework for DG is to extract the most informative and transferable underlying common features from source instances generated from different distributions and to generalize to unseen one. 
This kind of approach holds with the assumption that there exists an underlying invariant feature distribution among all domains, and that consequently such invariant features can generalize well to a target domain.~\cite{muandet2013domain} implemented MMD as
a distribution regularizer and proposed the kernel-based \emph{Domain Invariant Component Analysis} (DICA) algorithm.
An autoencoder-based model was proposed by~\cite{ghifary2015domain} under a multi-task learning setting to learn domain-invariant features via adversarial training.
\cite{li2018deep} proposed an end-to-end deep domain generalization approach by leveraging deep neural networks for domain-invariant representation learning.
\cite{Motiian_2017_ICCV} proposed to minimize the semantic alignment loss as well as the separation loss based on deep learning models. 
\cite{li2018domain} proposed a low-rank Convolutional Neural Network model based on domain shift-robust deep learning methods.

 There are also some approaches to tackle the domain generalization problems in a meta-learning manner. To the best of our knowledge,~\cite{li2018learning} first proposed to adopt the Meta Agnostic Meta-Learning (MAML)~\citep{finn2017model} which back-propagates the gradients of ordinary loss function of meta-test tasks. As pointed by~\cite{dou2019domain}, such an approach might lead to a sub-optimal solution, as it is highly abstracted from the feature representations. ~\cite{balaji2018metareg} proposed \emph{MetaReg} algorithm in which a regularization function ($e.g.$ weighted $L_1$ loss) is implemented for the classification layer of the model but not for the feature extractor layers. Then,~\citep{li2019feature} proposes an auxiliary meta loss which is gained based on the feature extractor. Furthermore, the network architecture of~\citep{li2019feature} is the widely used feature-critic style model based on a similar model from domain adversarial training technique~\citep{ganin2016domain}.~\cite{dou2019domain} and~\cite{dg_mmld} also started to implement clustering techniques on the invariant feature space for better classification and showed better performance on the target domain.

\subsection{Metric Learning}
\label{sect:related_works_metric_learning}
Metric learning aims to learn a discriminative feature embedding where similar samples are closer while different samples are further apart~\citep{8964455}.
\cite{hadsell2006dimensionality} proposed the \emph{siamese network} together with \emph{contrastive loss} to guide the instances stay close with each other in the feature space if they have the same labels and push them apart vice-versa.  
\cite{schroff2015facenet} proposed the \emph{triplet loss} aiming to learn a feature space where a positive pair has higher similarity than the negative pair when comparing by the same anchor with a given margin. 
\cite{oh2016deep} showed that neither the \emph{contrastive loss} nor  \emph{triplet loss} could efficiently explore the full pair-wise relations between instances under the mini-batch training setting. They further propose the \emph{lifted structure loss} to fully utilize pair-wise relations across batches. However, it only choose equal number of positive pairs as negative ones randomly, and many informative pairs are discarded~\citep{wang2019multi}, which restricts the ability of finding the informative pairs.
~\cite{yi2014deep} proposed the binomial deviance loss which could measure the hard pairs.
One remarkable work by~\cite{wang2019multi} combines the advantages both from~\emph{lifted structure loss} and \emph{binomial loss} to leverage the pair-similarity. 
They proposed to leverage not only pair-similarities (positive or negative pairs with each other) but also self-similarity which enables the learner to collect and weight informative pairs (positive or negative pairs) under an iterative (mining and weighting) manner. For a pair of instances, the self-similarity is gained from itself. 
Such a multi-similarity has been shown could measure the similarity and could cluster the samplers more efficiently and accurately. 
In the context of domain generalization, ~\cite{dou2019domain} proposed to guide the learner to leverage from the local similarity in the semantic feature space, in which the authors argued may contain essential domain-independent \emph{general knowledge} for domain generalization and adopt the constrative loss and triplet loss to encourage the clustering for solving this issue. Leveraging from the across-domain class similarity information can encourage the learner to extract robust semantic features that regardless of domains, which is an useful auxiliary information for the learner. If the learner could not separate the samples (from different source domains) with domain-independent class-specific cohesion and separation on the domain invariant feature space, it would still suffer from ambiguous decision boundaries. This ambiguous decision boundaries might still be sensitive to the unseen target domain.~\cite{dg_mmld} implement unsupervised clustering on source domains and showed better classification performance. Our work is orthogonal to previous works, proposing to enforce more distinguishable invariant features space via Wasserstein adversarial training and encouraging to leverage from label similarity information for better classification boundary.

\begin{table}[]\caption{{List of notations}}
\label{Neurocom_DG_list_of_notation}
\resizebox{0.93\textwidth}{!}{{
\begin{tabular}{@{}cl|cl@{}}
\toprule
\textbf{\textbf{Symbol}}           & \multicolumn{1}{c|}{\textbf{\textbf{Meaning}}}                                                                                & \textbf{\textbf{Symbol}} & \multicolumn{1}{c}{\textbf{\textbf{Meaning}}}                                                                                                                                                                 \\ \midrule
$F$                                & The feature extraction function                                                                                               & $\btheta_f$               & Parameter of feature extraction network                                                                                                                                                                       \\
$D$                                & The critic function                                                                                                           & $\btheta_d$               & Parameter of critic network                                                                                                                                                                                   \\
$C$                                & The classification function                                                                                                   & $\btheta_c$               & Parameter for classification network                                                                                                                                                                          \\
$m$                                & The number of source domains                                                                                                  & $\mathbf{x}_{j}^{(i)}$   & The $i$-th instance from the $j$-th domain                                                                                                                                                                    \\
$N_i$                              & \begin{tabular}[c]{@{}l@{}}The number of instances\\  in the $i$-th domain\end{tabular}                                       & $\mathbf{X}^{(i)}$       & \begin{tabular}[c]{@{}l@{}}The set of instances in the $i$-th domain\\ $\mathbf{X}^{(i)}=\{\mathbf{x}_{j}^{(i)}\}_{j=1}^N$\end{tabular}                                                                       \\
$\mathcal{D}$                      & \begin{tabular}[c]{@{}l@{}}The data distribution.\\ $\mathcal{D}_i$ are the source \\ domain distributions\end{tabular}       & $Z^{(i)}$                & The extracted feature from domain $i$                                                                                                                                                                         \\
$W_1(\mathcal{D}_i,\mathcal{D}_j)$ & \begin{tabular}[c]{@{}l@{}}Wasserstein-1 distance over two \\  distributions $\mathcal{D}_i$ and $\mathcal{D}_j$\end{tabular} & $y_j$                    & The label for corresponding instance $x_j$                                                                                                                                                                    \\
$\mathbf{S}$                       & The similarity matrix                                                                                                         & $S_{i,j}$                & \begin{tabular}[c]{@{}l@{}}The value of $i$-th row and $j$-th column \\ of the similarity matrix $\mathbf{S}$\end{tabular}                                                                                    \\
$w_{i,j}$                          & The weight for similarity $S_{i,j}$                                                                                           & $\epsilon$               & \begin{tabular}[c]{@{}l@{}}Small margin for roughly select \\ the positive and negative pairs\end{tabular}                                                                                                    \\
$\alpha$                           & \begin{tabular}[c]{@{}l@{}}Fixed parameter for \\ positive mining\end{tabular}                                                & $\beta$                  & Fixed parameter for negative mining                                                                                                                                                                           \\
$\lambda$                          & \begin{tabular}[c]{@{}l@{}}Parameter for \\ self-similarity mining\end{tabular}                                               & $\lambda_d$              & \begin{tabular}[c]{@{}l@{}}Coefficient for regularizing\\  the adversarial objective\end{tabular}                                                                                                             \\
$\lambda_s$                        & \begin{tabular}[c]{@{}l@{}}Coefficient for regularizing\\  the metric learning objective\end{tabular}                         & $\mathcal{L}$            & \begin{tabular}[c]{@{}l@{}}The objective functions, \\ $\mathcal{L}_C$ is the classification loss,\\ $\mathcal{L}_D$ is the adversarial loss,\\ $\mathcal{L}_{MS}$ is the metric similarity loss\end{tabular} \\ \bottomrule
\end{tabular}}}
\end{table}

\section{Preliminaries and Problem Setup}
We start by introducing some preliminaries. {In order to better summarize the notations symbols in this work, we provide the list of notations and symbols in Table~\ref{Neurocom_DG_list_of_notation}. }

\subsection{Notations and Definitions}
Following~\cite{redko2017theoretical} and \cite{Li2017dg}, suppose we have $m$ known source domains distributions $\{\mathcal{D}_i\}_{i=1}^m$, and $i^{th}$ domain contains $N_i$ labeled instances in total, denoted by $\{(\mathbf{x}^{(i)}_j,y^{(i)}_j)\}^{N_i}_{j=1}$, where $\mathbf{x}^{(i)}_j\in\mathbb{R}^{n}$ is the $j^{th}$ instance feature from the $i^{th}$ domain and $y^{(i)}_j\in \{1,\dots,K\}$ are the corresponding labels. 
For a hypothesis class $\mathcal{H}$, the expected source and target risk of a hypothesis $h\in\mathcal{H}$ over domain distribution $\mathcal{D}_i$ is the probabilities that $h$ wrongly predicts on the entire distribution $\mathcal{D}_i$: 
        $\epsilon_i(h)=\mathbb{E}_{(\mathbf{x},y)\sim\mathcal{D}_i}\ell(h(\mathbf{x},y))$,
where $\ell(\cdot)$ is the loss function. The empirical loss is also defined by: $\hat{\epsilon}_i(h)=\frac{1}{N_i}\sum_{j=1}^{N_i}\ell(h(\mathbf{x}_j,y_j))$.

In the setting of domain generalization, we only have the access to the seen source domains $\mathcal{D}_i$ but have no information about the target domain. The learner is expected to extract the underlying invariant feature space across the source domains and generalize to a new target domain.

\subsection{Optimal Transport and Wasserstein Distance}

We follow \cite{redko2017theoretical} and define $c:\mathbb{R}^n\times\mathbb{R}^n \to \R^{+}$ as the cost function for transporting one unit of mass $\x$ to $\x'$, then the primal form of the Wasserstein distance between $\mathcal{D}_i$ and $\mathcal{D}_j$ could be computed by,
\begin{equation}
W_p^p(\mathcal{D}_i,\mathcal{D}_j) = \inf_{\gamma\in \Pi(\mathcal{D}_i,\mathcal{D}_j)} \int_{\mathbb{R}^n\times\mathbb{R}^n}c(\mathbf{x},\mathbf{x}^\prime)^p d\gamma(\mathbf{x},\mathbf{x}^\prime)
\label{original_eq_wasserstein}
\end{equation}
where $\Pi(\mathcal{D}_i,\mathcal{D}_j)$ is the probability coupling on $\mathbb{R}^n\times\mathbb{R}^n$ with marginals $\mathcal{D}_i$ and $\mathcal{D}_j$ referring to all the possible coupling functions. Throughout this paper, we adopt Wasserstein-1 distance only ($p=1$). 

{Computing the primal form of Wasserstein distance (Eq.~\ref{original_eq_wasserstein}) is computational inefficiently. Assuming $|\mathcal{D}_i|=n, |\mathcal{D}_j|=m$, the time complexity for directly computing Eq.~\ref{original_eq_wasserstein} is $\mathcal{O}(n^3+m^3)$. On the contrary, leveraging the \emph{Kantorovich-Rubinstein duality}~\citep{wainwright2019high} of Wasserstein distances could help to get a more efficient approximation. Assume $f$ a $1$-Lipschitz-continuous $w.r.t.$ the cost function: $\|f(x)-f(x^\prime)\| \leq c(x,x^\prime)$, we can prove that for any function $f$,
\begin{equation*}
    W_1 (\mathcal{D}_i,\mathcal{D}_j)  \geq \mathbb{E}_{x\sim \mathcal{D}_i} d(x) - \mathbb{E}_{x^\prime \sim \mathcal{D}_j} d(x^\prime)
\end{equation*}
The equality arrives when $f$ reaches the maximum of the right side,}

\begin{equation}
\label{dual_original_eq_wasserstein}
    W_1(\mathcal{D}_i,\mathcal{D}_j) = \sup_{\|f\|_L < 1} \mathbb{E}_{x\in\mathcal{D}_i}f(x) - \mathbb{E}_{x^\prime \in \mathcal{D}_j}f(x^\prime)
\end{equation}

{In practice, such a function $f$ could be approximated by a neural-network, which allows us to compute this Kantorovich-Rubinstein duality efficiently by computing the expectation and the complexity $w.r.t.$ $f(x)$ is only $\mathcal{O}(n+m)$. Empirically, to compute the $\sup$ is equivalent to find out the maximum of $W_1$ (by an $\arg \max$ operation). General neural network optimizer ($e.g.$ SGD or Adam) can efficiently solve the maximum problem to evaluate the dual value of $W_1$ distance.}
\begin{figure}
		\centering 
\includegraphics[width=0.60\textwidth]{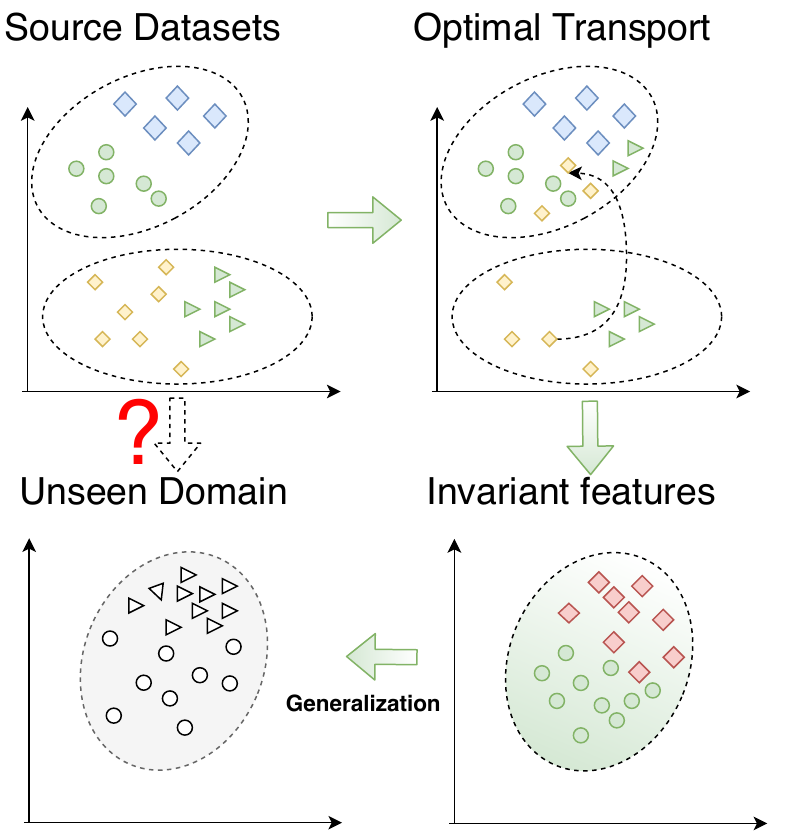}
\caption{Use optimal transport (OT) for domain generalization: Typically to directly predict on the unseen domain (the white dashed arrow) is difficult. In order to learn domain invariant features, as showed in the direction of the green arrow we adopted the OT technique to achieve domain alignments for extracting invariant features. After the OT transition, the invariant features can be generalized to unseen domain.} \label{fig_ot_dg}
\end{figure}

Optimal transport theory and Wasserstein distance were recently investigated in the context of machine learning \citep{arjovsky2017wasserstein} especially in the domain adaptation area \citep{courty2016optimal,zhou2020discriminative}. The general idea of implementing the optimal transport technique for domain generalization across domains is illustrated in Fig.~\ref{fig_ot_dg}. To learn domain invariant features, OT technique is implemented to achieve domain alignments for extracting invariant features. After the OT transition, the invariant features can be generalized to unseen domain. 

\subsection{Metric Learning}
 For a pair of instances $(\x_i,y_i)$ and $(\x_j,y_j)$, the notion of \emph{positive pairs} usually refers to the condition where pair $i,j$ have same labels ($y_i=y_j$), while the negative pairs usually refers to the condition $y_i\neq y_j$. The central idea of metric learning is to encourage a pair of instances who have the same labels to be closer, and push negative pairs to be apart from each other~\citep{wu2017sampling}.

Follow the framework of~\cite{wang2019multi}, we show the general pair-weighting process of metric learning. Assuming the feature extractor $f$ parameterized by $\btheta_f$ projects the instance $\mathbf{x} \in\mathbb{R}^{n}$ to a $d$-dimensional normalized space: $f(\mathbf{x};\btheta_f):\; \mathbb{R}^n\to [0,1]^d$. Then, for two samples $\mathbf{x}_i$ and $\mathbf{x}_j$, the similarity between them could be defined as the inner product of the corresponding feature vector:
\begin{equation}
    S_{i,j}:=\langle f(\mathbf{x}_i;\btheta_f),f(\mathbf{x}_j;\btheta_f)\rangle
    \label{Eq.dotsimilarity}
\end{equation}
To leverage the across-domain class similarity information can encourage the learner to extract the classification boundary that regardless of domains, which is an useful auxiliary information for the learner. We further elaborate it in section~\ref{metric_learning_domain_agnostic}.

\section{Proposed Method}
\begin{figure}
\centering
\begin{subfigure}{.535\textwidth}
  \centering
  % include first image
  \includegraphics[width=\textwidth]{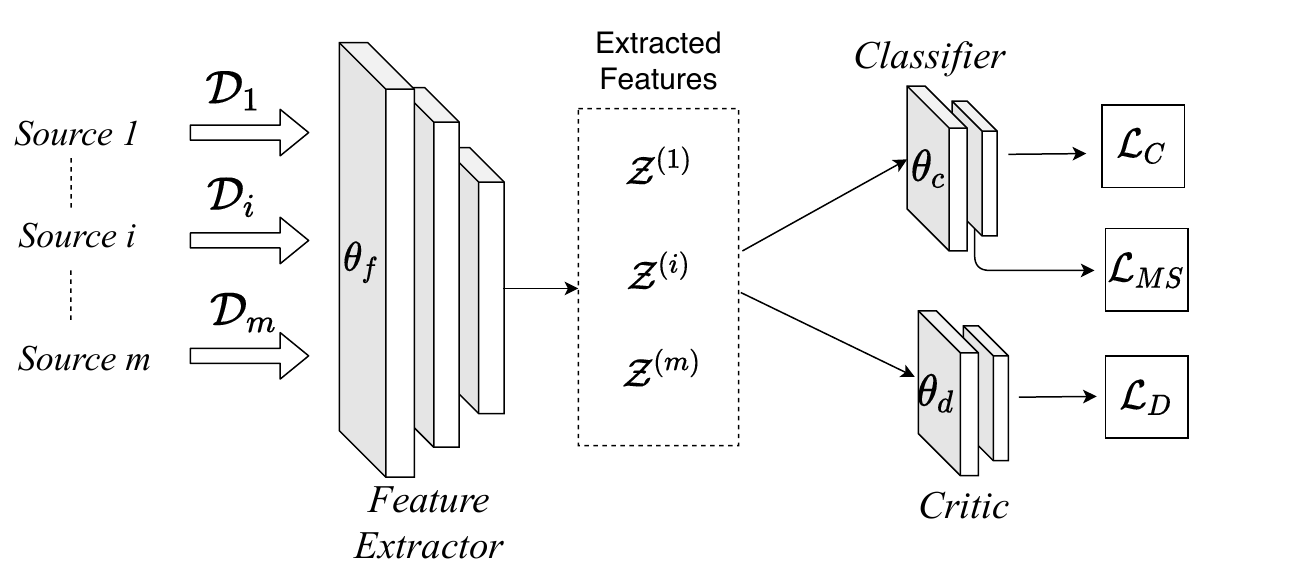}  
  \caption{The whole workflow the proposed WADG model.}
  \label{fig:sub-general_workflow}
\end{subfigure}
\;
\begin{subfigure}{.43\textwidth}
  \centering  % include second image
  \includegraphics[width=\textwidth]{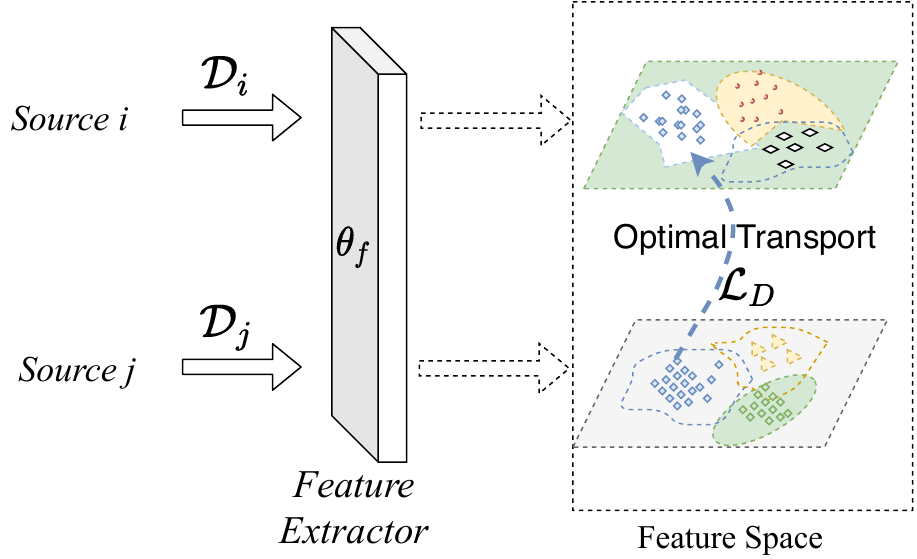}  
  \caption{Optimal Transport for Feature Alignment.}
  \label{fig:sub-OT alignment}
\end{subfigure}\\
\begin{subfigure}{1.\textwidth}
  \centering
  % include third image
  \includegraphics[width=\textwidth]{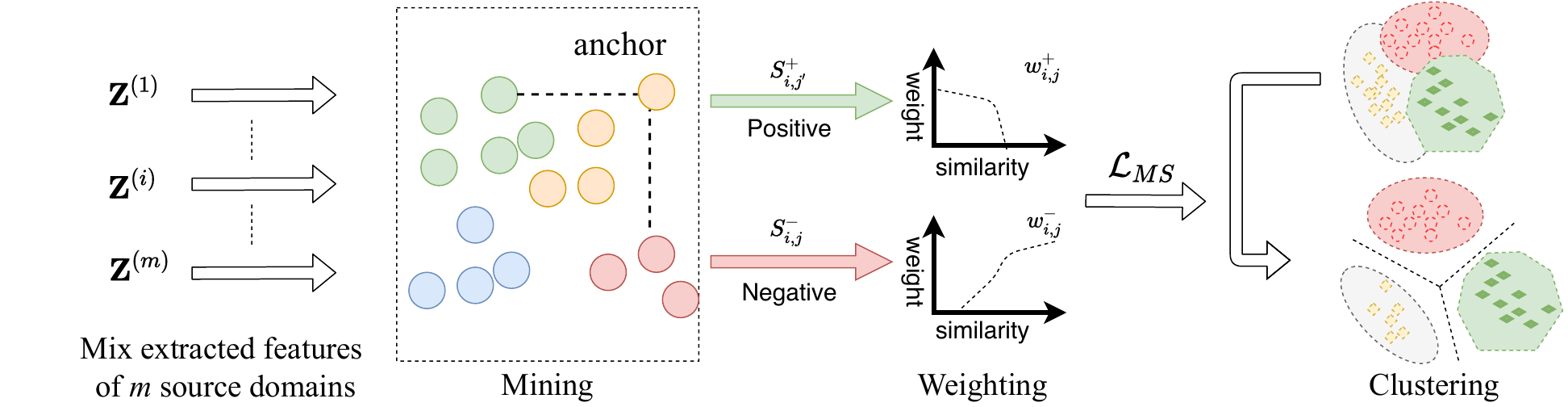}  
  \caption{Metric Learning for Clustering Proces}
  \label{fig:sub-metric learning}
\end{subfigure}

\caption{The proposed WADG method. (a): the general workflow of WADG method. The model mainly consists of three parts, the feature extractor, classifier and critic function. During training, the model receives all the source domains. The feature extractor is trained to learn invariant features together with the critic function in an adversarial manner. (b): For each pair of source domains $\mathcal{D}_i$ and $\mathcal{D}_j$, optimal transport process for aligning the features from different domains. (c): The metric learning process. For a batch of all source domain instances, we first roughly mining the positive and negative pairs via Eq.~\ref{eq_round_mining}. Then, compute the corresponding weights via Eq.~\ref{Eq.negative_pair_weight} and Eq.~\ref{Eq.positive_pair_weight} to compute $\mathcal{L}_{MS}$ to guide the clustering process.}
\label{fig:whole workflow}
\end{figure}

The high-level idea of WADG algorithm is to learn a domain-invariant feature space and domain-agnostic classification boundary. Firstly, we align the marginal distribution of different source domains via optimal transport by minimizing the Wasserstein distance to achieve the domain-invariant feature space. And then, we adopt metric learning objective to guide the learner to leverage the class similarity information for a better classification boundary. A general workflow of our method is illustrated in Fig.~\ref{fig:sub-general_workflow}. The model contains three major parts: a feature extractor, a classifier and a critic function. 

The feature extractor function $F$, parameterized by $\btheta_f$, extracts the features from different source domain. For set of instances $\mathbf{X}^{(i)}=\{\x_{j}^{(i)}\}_{j=1}^{N_i}$ from domain $\mathcal{D}_i$, we can then denote the extracted feature from domain $i$ as $\mathbf{Z}^{(i)}=F(\mathbf{X}^{(i)})$. The classification function $C$, parameterized by $\btheta_c$, is expected to learn to predict labels of instances from all the domains correctly. The critic function $D$, parameterized by $\btheta_d$, aims to measure the empirical Wasserstein distance between features from a pair of source domains. For the target domain, all the instances and labels are absent during the training time. 

WADG aims to learn the domain-agnostic features with distinguishable classification boundary. During each train round, the network receives the labelled data from all domains and train the classifier under a supervised mode with the classification loss $\mathcal{L}_C$. For the classification process, we use the typical cross-entropy loss for all $m$ source domains:
 \begin{equation}
    \mathcal{L}_{C} = -\sum_{i=1}^m\sum_{j=1}^{N_i} y_j\log(\mathbb{P}(C(F(\textbf{x}_j^{(i)}))))  
    \label{Eq.cls_loss_all_domains}
 \end{equation}

Through this, the model could learn to train the category information on over all the domains. 
The feature extractor $F$ is then trained to minimize the estimated Wasserstein Distance in an adversarial manner with the critic $D$ with an objective $\mathcal{L}_D$. We then adopt a metric learning objective (namely, $\mathcal{L}_{MS}$) for leveraging the similarities for a better classification boundary. Our full method then solve the joint loss function,
\begin{equation*}
\mathcal{L} = \arg\min_{\theta_f,\theta_c}\max_{\theta_d} \mathcal{L}_{C}+\mathcal{L}_{D} + \mathcal{L}_{MS},
\end{equation*}
where $\mathcal{L}_D$ is the adversarial objective function, and $\mathcal{L}_{MS}$ is the metric learning objective function. In the sequel, we will elaborate these two objectives in section~\ref{wasserstien_over_all_domains} and section~\ref{metric_learning_domain_agnostic}, respectively.

\subsection{Adversarial Domain Generalization via Optimal Transport}
\label{wasserstien_over_all_domains}

As optimal transport could constrain labelled source samples of the same class to remain close during the transportation process~\citep{courty2016optimal}. We deploy optimal transport with Wasserstein distance~\citep{redko2017theoretical,shen2017wasserstein} for aligning the marginal feature distribution over all the source domains. 

A brief workflow of the optimal transport for a pair of sourcce domains is illustrated in Fig.~\ref{fig:sub-OT alignment}. The critic function $D$ estimates the empirical Wasserstein Distance between the each source domain through a pair of instances from the empirical sets $\x^{(i)}\in\mathbf{X}^{(i)}$ and $\x^{(j)}\in\mathbf{X}^{(j)}$. In practice~\citep{shen2017wasserstein}, the dual term Eq.~\ref{dual_original_eq_wasserstein} of Wasserstein distance could be computed by,
\begin{equation}
    \begin{split}
            W_1(\mathbf{X}^{(i)}, \mathbf{X}^{(j)})=\max \big( \frac{1}{N_{i}}\sum_{\x^{(i)}\in\mathbf{X}^{(i)}}D(F(\mathbf{x}^{(i)}))         -\frac{1}{N_{j}}\sum_{\mathbf{x}^{(j)}\in\mathbf{X}^{(j)}}D(F(\x^{(j)})) \big )
            \label{Eq.Wasserstein}
    \end{split}
\end{equation}

As in domain generalization setting, there usually exists more that two source domains, we can sum all the empirical Wasserstein distance between each pair of source domains,
\begin{equation}
    \mathcal{L}_D = \sum_{i=1}^m\sum_{j=i+1}^m\big [ \frac{1}{N_i}\sum_{\x^{(i)}\in\mathbf{X}^{(i)}}D(F(\x^{(i)}))-\frac{1}{N_j}\sum_{\x^{(j)}\in\mathbf{X}^{(j)}}D(F(\x^{(j)}))\big]
    \label{Eq.mult_domain_Wa}
\end{equation}

Throughout this pair-wise optimal transport process, the learner could extract a domain-invariant feature space, we then propose to apply metric learning approaches to leverage the class label similarity for domain independent clustering feature extraction. We then introduce the metric learning for domain agnostic clustering in the next section.

\subsection{Metric Learning for Domain Agnostic Classification Boundary}
\label{metric_learning_domain_agnostic}

As aforementioned, only aligning the marginal features via adversarial training is not sufficient for DG since there may exist a ambiguous decision boundary~\citep{dou2019domain}. When predicting on the target domain, the learner may still suffer from this ambiguous decision boundary. To this end, we adopt the metric learning techniques~\citep{wang2019multi} to help cluster the instances and promote a better prediction boundary for better generalization.

To solve this, except to the supervised source classification and alignment of the marginal distribution across domains with the Wasserstein adversarial training defined above, we then further encourage robust domain-independent local clustering via leverage from label information using the metric learning objective. The brief workflow is illustrated in Fig.~\ref{fig:sub-metric learning}. Specifically, we adopt the metric learning objective to require the images regardless of their domains to follow the two aspects: 1) images from the same class are semantically similar, thereby should be mapped nearby in the embedding space (semantic clustering), while 2) instances from different classes should be mapped apart from each other in embedding space. Since goal of domain generalization aims to learn to hypothesis could predict well on all the domains, the clustering should also be achieved under a domain-agnostic manner.

To this end, we mix the instances from all the source domains together and encourage the clustering for domain agnostic features via the metric learning techniques to achieve a domain-independent clustering decision boundary. For this, during each training iteration, for a batch $\{\mathbf{x}_1^{(i)},y_1^{(i)},\dots,\mathbf{x}^{(i)}_b,y^{(i)}_b\}_{i=1}^m$ from $m$ source domains with batch size $b$, we mix all the instances from each domain and denoted by $\{(\mathbf{x}_i^B,y_i^B)\}_{i=1}^{m^\prime}$ with total size $m^\prime$. We first measure the relative similarity between the negative and positive pairs, which is introduced in the next sub-section.

\subsubsection{Pair Similarity Mining}
 Assume $\mathbf{x}_i^{B}$ is an anchor, a negative pair $\{\mathbf{x}_i^B,\mathbf{x}_j^B \}$ and a positive pair $\{\mathbf{x}_i^B,\mathbf{x}_{j^\prime}^B \}$ are selected if $S_{ij}$ and $S_{i,{j^\prime}}$ satisfy the negative condition $S_{i,j}^{-}$ and the positive condition $S_{i,j}^{+}$, respectively :
\begin{equation}
\label{eq_round_mining}
    S_{i,j}^{-}\ge  \min_{y_i= y_k} S_{i,k}-\epsilon, \; \; \; \; S_{i,j^\prime}^{+}\leq \min_{y_i\neq y_k} S_{i,k}+\epsilon
\end{equation}
where $\epsilon$ is a given margin. Through Eq.~\ref{eq_round_mining} and specific margin $\epsilon$, we will have a set of negative pairs $\mathcal{N}$ and a set of positive pairs $\mathcal{P}$. This process (Eq.~\ref{eq_round_mining}) could roughly cluster the instances with each anchor by selecting informative pairs (inside of the margin), and discard the less informative ones (outside of the margin).

With such roughly selected informative pairs $\mathcal{N}$ and $\mathcal{P}$, we then assign the instance with different weights. Intuitively, if a instance has higher similarity with an anchor, then it should stay closer with the anchor and vice-versa. We introduce the weighting process in the next section.

\subsubsection{Pair Weighting}~\label{sect:pair_weighting}

For instances of positive pairs, if they are more similar with the anchor, then it should have higher weights while give the negative pairs with lower weights if they are more dissimilar, no matter which domain they come from. Through this process, we can push the instances into several groups via measure their similarities.

 For $N$ instances, computing the similarity between each pair could result in a similarity matrix $\mathbf{S}\in\mathbb{R}^{N\times N}$.
For a loss function based on pair similarity, it can usually be defined by $\mathcal{F}(\mathbf{S},y)$. Let $S_{i,j}$ be the $i^{th}$ row, $j^{th}$ column element of matrix $\mathbf{S}$. The gradient $w.r.t$ the network could be computed by,

\begin{equation}
\begin{split}
        \frac{\partial\mathcal{F}(\mathbf{S},y)}{\partial\btheta_f}&= \frac{\partial\mathcal{F}(\mathbf{S},y)}{\partial\mathbf{S}}\frac{\partial \mathbf{S}}{\partial\btheta_f}=\sum_{i=1}^N\sum_{j=1}^N\frac{\partial\mathcal{F}(\mathbf{S},y)}{\partial S_{i,j}}\frac{\partial S_{i,j}}{\partial \btheta_f}
\end{split}
\label{Eq.dev_theta_f}
\end{equation}

 Eq.~\ref{Eq.dev_theta_f} could be reformulated into a new loss function $\mathcal{L}_{MS}$ as,
\begin{equation}
    \mathcal{L}_{MS}= \sum_{i=1}^N\sum_{j=1}^N\frac{\partial\mathcal{F}(\mathbf{S},y)}{\partial S_{i,j}}S_{i,j}
    \label{Eq.reformed_lms}
\end{equation}

 usually the metric loss defined $w.r.t$ similarity matrix $\mathbf{S}$ and label $y$ could be reformulated by Eq.~\ref{Eq.reformed_lms}. The term $\frac{\partial\mathcal{F}(\mathbf{S},y)}{\partial S_{i,j}}$ in Eq.~\ref{Eq.reformed_lms} could be treated as an constant scalar since it doesn't contain the gradient of $\mathcal{L}_{MS}$ $w.r.t$ $\btheta_f$. Then, we just need to compute the gradient term $\frac{\partial \mathcal{F}_{i,j}}{\partial \btheta_f}$ for the positive and negative pairs. Since the goal is to encourage the positive pairs to be closer, then we can assume the gradient $\leq 0$, $i.e.$, $\frac{\partial \mathcal{F}_{i,j}}{\partial \btheta_f}\leq 0$. Conversely, for a negative pair, we could assume $\frac{\partial \mathcal{F}_{i,j}}{\partial \btheta_f}\geq 0$. 
Thus, Eq.~\ref{Eq.reformed_lms} is transformed by the summation over all the positive pair ($y_i=y_j$) and negative pairs ($y_i\neq y_j$),
\begin{equation}
\begin{split}
        \mathcal{L}_{MS}&=\sum_{i=1}^N\sum_{j=1}^N\frac{\partial\mathcal{F}(\mathbf{S},y)}{\partial S_{i,j}}S_{i,j}\\
        &=\sum_{i=1}^N\left(\sum_{j=1,y_j\neq y_i}^N \frac{\partial\mathcal{F}(\mathbf{S},y)}{\partial S_{i,j}} S_{i,j}+\sum_{j=1,y_j= y_i}^N \frac{\partial\mathcal{F}\big(\mathbf{S},y)}{\partial S_{i,j}} S_{i,j} \right)\\
        &=\sum_{i=1}^N \left(\sum_{j=1,y_j\neq y_i}^N w_{i,j}S_{i,j} - \sum_{j=1,y_j= y_i}^N w_{i,j} S_{i,j}  \right)
\end{split}
    \label{L_ms_with_weight}
\end{equation}
where $w_{i,j}= \big | \frac{\partial S_{i,j}}{\partial \btheta_f} \big|$ is regarded as the weight for similarity $S_{i,j}$. 
Since our goal is to encourage the positive pairs to be closer, then we can assume the weight for positive pairs is smaller than 0. Conversely, for a negative pair, we can assume the weight is larger than 0. The intuition is that for a negative pair of instances, let the weight be positive, we can give it a higher loss value. Then, the learner can learn to distinguish them. On the contrary, we can assign the negative weights towards the positive pairs, which will guide the learner to not separate them apart. 
For each pair of instances $i,j$, we could assign different weights according to their similarities $S_{i,j}$. 
Then, we can denote $w_{i,j}^+$ and $w_{i,j}^-$ as the weight of a positive or negative pairs' similarity, respectively. 

Previously,~\cite{yi2014deep} and \cite{wang2019multi} applied a soft function for measuring the similarity. We then consider the similarity of the pair itself ($i.e.$ self-similarity), the negative similarity and the positive similarity. 
The weight of self-similarity could be measured by $\exp({S_{i,j}-\lambda})$ with a small threshold $\lambda$. For a selected negative pair $\{\x^B_i, \x^B_j\}\in \mathcal{N}$ the corresponding weight (see  Eq.~\ref{L_ms_with_weight}) could be defined by the soft function of self-similarity together with the negative similarity:
\begin{equation}
\begin{split}
    w_{i,j}^-&=\frac{1}{\exp(\beta(\lambda-S_{ij}))+\sum_{k\in\mathcal{N}}\exp(\beta(S_{i,k}-\lambda))}\\
        &= \frac{\exp(\beta(S_{ij}-\lambda))}{1+\sum_{k\in\mathcal{N}}\exp(\beta(S_{ik}-\lambda))}
\end{split}
    \label{Eq.negative_pair_weight}
\end{equation}

Similarly, the weight of a positive pair $\{\mathbf{x}^B_i,\mathbf{x}^B_j\}\in\mathcal{P}$ is defined by,

\begin{equation}
   w_{i,j}^+=\frac{1}{\exp(-\alpha(\lambda-S_{i,j}))+\sum_{k\in\mathcal{P}}\exp(-\alpha(S_{i,k}-S_{i,j}))}
    \label{Eq.positive_pair_weight}
\end{equation}

Then, take Eq.~\ref{Eq.negative_pair_weight} and Eq.~\ref{Eq.positive_pair_weight} into Eq.~\ref{L_ms_with_weight}, and integrate Eq.~\ref{L_ms_with_weight} with the similarity mining $S_{i,j}$, we have the objective function for clustering,
\begin{equation}
\begin{split}
       \mathcal{L}_{MS}=  \frac{1}{m}\sum_{i=1}^{m}\big\{\frac{1}{\alpha}\log[1+\sum_{k\in\mathcal{P}_i}\exp(-\alpha(S_{ik}-\lambda))] 
        +\frac{1}{\beta}\log[1+\sum_{k\in\mathcal{N}_i}\exp(\beta(S_{ik}-\lambda))]\big\} 
        \label{Eq.multi_similarity_loss}
\end{split}
\end{equation}
where $\lambda$, $\alpha$ and $\beta$ are fixed hyper-parameters, we elaborate them in the empirical setting section~\ref{sect:implementation_details}. Then, the whole objective of our proposed method is,
\begin{equation}
\mathcal{L} = \arg\min_{\theta_f,\theta_c}\max_{\theta_d} \mathcal{L}_{C}+\lambda_d\mathcal{L}_{D} +\lambda_s \mathcal{L}_{MS}
\label{Eq:full_objective}
\end{equation}
where $\lambda_d$ and $\lambda_s$ are coefficients to regularize $\mathcal{L}_d$ and $\mathcal{L}_{MS}$ respectively.

Based on these above, we propose the WADG algorithm in Algorithm~\ref{wadg_algo}. And we show the empirical results in the next section.

\begin{center}
\begin{algorithm}[t!]
		\caption{The proposed WADG algorithm (one round)}
		\begin{algorithmic}[1] 
		\REQUIRE Samples from different source domains $\{{\mathcal{D}}_i\}_{i=1}^M$
        \ENSURE Neural network parameters $\btheta_{f}$, $\btheta_c$, $\btheta_d$ 
        \FOR{mini-batch of samples $\{(\mathbf{x}^{(i)}_s,y^{(i)}_s)\}$ from source domains} 
        \STATE Compute the classification loss $\mathcal{L}_C$ over all the domains according to Eq.~\ref{Eq.cls_loss_all_domains}
        \STATE Compute the Wasserstein distance $\mathcal{L}_D$ between each pair of source domains according to Eq.~\ref{Eq.mult_domain_Wa}

        \STATE Mix the pairs from different domains and compute the similarity by Eq.~\ref{Eq.dotsimilarity}
        \STATE Roughly select the positive and negative pairs by solving Eq.~\ref{eq_round_mining}

        \STATE Compute similarity loss $\mathcal{L}_{MS}$ on all the source instances by Eq.~\ref{Eq.multi_similarity_loss}
        
        \STATE    Update $\theta_f,\theta_c$ and $\baselinestretch_d$ by solving Eq.~\ref{Eq:full_objective} with learning rate $\eta$:
           \begin{equation*}
           \begin{split}
               &\btheta_f \leftarrow \btheta_f - \eta\frac{\partial(\mathcal{L}_{C}+\lambda_d\mathcal{L}_D+\lambda_s\mathcal{L}_{MS})}{\partial\btheta_f}, \\
               &\btheta_c \leftarrow \btheta_c - \eta\frac{\partial(\mathcal{L}_{C}+\lambda_d\mathcal{L}_D+\lambda_s\mathcal{L}_{MS})}{\partial\btheta_c}, \\
               &\btheta_d \leftarrow \btheta_d + \eta\frac{\partial \mathcal{L}_D}{\partial\btheta_d}
           \end{split}
           \end{equation*}
 
        \ENDFOR
		\STATE Return the optimal parameters $\btheta_f^\star$, $\btheta_c^\star$ and $\btheta_d^\star$
        \end{algorithmic}
        \label{wadg_algo}
\end{algorithm}
\end{center}

\section{Experiments and Results}
\label{experiments results}
\subsection{Datasets}
In order to evaluate our proposed approach, we implement experiments on {three} common used datasest: \textbf{VLCS}~\citep{torralba2011unbiased}, \textbf{PACS}~\citep{Li2017dg} {and \textbf{Office-home}~\citep{venkateswara2017Deep}} dataset. The VLCS dataset contains images from 4 different domains: PASCAL VOC2007 (V), LabelMe (L), Caltech (C), and SUN09 (S). Each domain includes five classes: \emph{bird, car, chair, dog} and \emph{person}. PACS dataset is a recent benchmark dataset for domain generalization. It consists of four domains: Photo (P), Art painting (A), Cartoon (C), Sketch (S), with objects from seven classes: dog, elephant, giraffe, guitar, house, horse, person. {Office-Home is a more challenging dataset, which contains four different domains: \emph{Art} (Ar), \emph{Clipart} (Cl), \emph{Product} (Pr) and \emph{Real World} (Rw), with $65$ categories in each domain. Previous work showed that matter the adversarial model is trained under supervised~\citep{long2017learning}, semi-supervised~\citep{zhou2020discriminative} or unsupervised~\citep{long2018conditional} way, the model will suffer from learning the diverse feature. To test our domain generalization model on this dataset could also help to affirm the effectiveness of our approach.}

\subsection{Baselines and Implementation details}
\label{sect:implementation_details}
To show the effectiveness of our proposed approach, we compare our algorithm on the benchmark datasets with the following recent domain generalization methods. 

\begin{itemize}
    \item \textbf{\emph{Deep All}}: We follow the standard evaluation protocol of Domain Generalization to set up the pre-trained Alexnet or ResNet-18 fine-tuned on the aggregation of all source domains with only the classification loss.
  
    \item TF~\citep{li2017deeper}:  A low-rank parameterized Convolution Neural Network model which aims to reduce the total number of model parameters for an end-to-end Domain Generalization training. 
    \item CIDDG~\citep{li2018deep}: Matches the conditional distribution by change the class prior.
    \item MLDG~\citep{li2018learning}: The meta-learning approach for domain generalization. It runs the meta-optimization on simulated meta-train/ meta-test sets with domain shift
    \item CCSA~\citep{Motiian_2017_ICCV}: 
    The contrastive semantic alignment loss was adopted together with the source classification loss function for both the domain adaptation and domain generalization problem.
    \item MMD-AAE~\citep{li2018domain}: The Adversarial Autoencoder model was adopted together with the Mean-Max Discrepancy to extract a domain invariant feature for generalization.
    \item D-SAM~\citep{d2018domain}: It aggregates domain-specific modules and merges general and specific information together for generalization.
\item JiGen~\citep{carlucci2019domain}: It achieves domain generalization by solving the Jigsaw puzzle via the unsupervised task.
\item MASF~\citep{dou2019domain}: A meta-learning style method which based on MLDG and combined with Consitrastive Loss/ Triplet Loss to encourage domain-independent semantic feature space.
\item MMLD~\citep{dg_mmld}: An approach that mixes all the source domains by assigning a pseudo domain label for extract domain-independent cluster feature space.
    \end{itemize}

\begin{table}[]
\caption{The hyper-parameter values for experiments}
\centering\label{value of hyperparameters}
{\begin{tabular}{@{}cc|cc@{}}
\toprule
\textbf{Hyper-parameters}      & \textbf{Value}                         & \textbf{Hyper-parameters} & \textbf{Value} \\ \midrule
\multirow{2}{*}{learning rate} & PACS: $5\times 10^{-4}$                & $\lambda$                 & $1.0$          \\ \cmidrule(l){3-4} 
                               & Office-home: $2\times 10^{-4}$         & $\alpha$                  & $2.0$          \\ \midrule
$\lambda_d$                    & $\lambda_d= \frac{2}{1+\exp(-10 p)}-1$ & $\beta$                   & $40.0$         \\ \midrule
$\lambda_s$                    & $[1e-4,1e-5]$                          & $\epsilon$                & $0.1$          \\ \bottomrule
\end{tabular}}
\end{table}

Following the general evaluation protocol of domain generalization ($e.g.$~\cite{dou2019domain,dg_mmld}), on PACS and VLCS dataset. We first test our algorithm on by using AlexNet~\citep{krizhevsky2012imagenet} backbones by removing the last layer as feature extractor. For preparing the dataset, we follow the train/val./test split and the data pre-processing protocol of~\cite{dg_mmld}. As for the classifier, we initialize a three-layers MLP whose input has the same number of inputs as the feature extractor's output and to have the same number of outputs as the number of object categories (2048-256-256-$K$), where $K$ is the  number of classes. For the critic network, we also adopt a three-layers MLP (2048-1024-1024-1). For the metric learning objective, we use the output of the second layer of classifier network (with size 256) for computing the similarity.

{In order to better demonstrating the hyper-parameters used in this work, we firstly summarized the value of hyper-parameters in Table.~\ref{value of hyperparameters}. The corresponding descriptions are provided in the following parts.}
We adopt the ADAM~\citep{kingma2014adam} optimizer for training with learning rate ranging from $5\times 10^{-4}$ to $5\times 10^{-5}$ for the whole model together with mini-batch size $64$.\\
\begin{table}[]
\caption{Empirical Results (accuracy $\%$) on PACS dataset with pre-trained AlexNet as Feature Extractor. For each column, we refer the generalization taks as the target domain name. For example, the third column `Cartoon‘ refers to the generalization tasks where domain \emph{Cartoon} is the target domain while the model is trained on the rest three domains. }\label{tb_pacs_dataset}
\centering
\resizebox{1.0\textwidth}{!}{\begin{tabular}{l|llll|l}

\toprule
Method   &  Art   & Cartoon  & Sketch & Photo & Avg. \\
\hline
Deep All &  $63.30$ & $63.13$ & $54.07$ & $87.70$ & $67.05$\\
TF\citep{li2017deeper} & $62.86$ & $66.97$ & $57.51$ & $89.50$ & $59.21$  \\
CIDDG\citep{li2018deep}  & $62.70$ & $69.73$ & $64.45$ & $78.65$ & $68.88$ \\
MLDG \citep{li2018learning}&  $66.23$ & $66.88$ & $58.96$ & $88.00$ & $70.01$\\
D-SAM\citep{d2018domain}  & $63.87$ & $70.70$ & $64.66$ & $85.55$ & $71.20$\\
JiGen\citep{carlucci2019domain}        &  $67.63$ & $71.71$ & $65.18$ & $89.00$ & $73.38$\\
MASF\citep{dou2019domain}  &  $\mathbf{70.35}$ & $ 72.46$  & $67.33$ &$90.68$ & $75.21$\\
MMLD\citep{dg_mmld}        &  $69.27$ & $\mathbf{72.83}$  & $66.44$ &$88.98$ & $74.38$\\

\hline
Ours &$70.21$ & $72.51$ & $\mathbf{70.32}$ & $\mathbf{89.81}$ & $\mathbf{75.71}$ \\
\bottomrule
\end{tabular}}
\end{table}

{ For stable training, we set coefficient $\lambda_d= \frac{2}{1+\exp(-10 p)}-1$ to regularize the adversarial loss, where $p$ is the training progress, to regularize the adversarial loss. This regularization scheme $\lambda_d$ has been widely used in adversarial training based domain adaptation and generalization setting ($e.g.$~\citep{long2017learning,wen2019bayesian,dg_mmld}) and have been proved could help to stabilize the training process.} For the setting of $\lambda_s$, we follow the setting of~\citep{dou2019domain} and set the value to $10^{-4}$. {In our preliminary validation results, the performance is not sensitive with $\lambda_d \in [0,1]$. We also tried to range $\lambda_s$ from $10^{-3}$ to $10^{-6}$ via reverse validation and didn't observe obvious differences.}  \\

\begin{table}[]

\resizebox{1.0\textwidth}{!}{\begin{tabular}{l|llll|l}
\toprule
Method   & Caltech  & LabelMe & Pascal  & Sun     & Avg.    \\ \midrule
Deep All & $92.86$ & $63.10$ & $68.67$ & $64.11$ & $72.19$ \\
D-MATE~\citep{ghifary2015domain}   & $89.05$  & $60.13$ & $63.90$ & $61.33$ & $68.60$ \\
CIDDG~\citep{li2018deep}     & $88.83$  & $63.06$ & $64.38$ & $62.10$ & $69.59$ \\
CCSA~\citep{Motiian_2017_ICCV}    & $92.30$  & $62.10$ & $67.10$ & $59.10$ & $70.15$ \\
SLRC~\citep{ding2017deep}     & $92.76$  & $62.34$ & $65.25$ & $63.54$ & $70.97$ \\
TF~\citep{li2017deeper}     & $93.63$  & $63.49$ & $69.99$ & $61.32$ & $72.11$ \\
MMD-AAE~\citep{li2018domain}  & $94.40$  & $62.60$ & $67.70$ & $64.40$ & $72.28$ \\
D-SAM~\citep{d2018domain}    & $91.75$  & $56.95$ & $58.95$ & $60.84$ & $67.03$ \\
MLDG~\citep{li2018learning}     & $94.4$   & $61.3$  & $67.7$  & $65.9$  & $73.30$ \\
JiGen~\citep{carlucci2019domain}    & $96.93$  & $60.90$ & $70.62$ & $64.30$ & $73.19$ \\
MASF~\citep{dou2019domain}  & $94.78$  & $\mathbf{64.90}$ & $69.14$ & $67.64$ & $74.11$ \\
MMLD~\citep{dg_mmld}  & $96.66$  & $58.77$ & $\mathbf{71.96}$ & $\mathbf{68.13}$ & $73.88$ \\ \hline
Ours     & $\mathbf{96.68}$   & $64.26$    & $71.47$     &  $66.62$   &  $\mathbf{74.76}$  \\ \bottomrule
\end{tabular}}
\caption[Empirical Results (accuracy $\%$) on VLCS dataset with pre-trained AlexNet as feature extractor.]{Empirical Results (accuracy $\%$) on VLCS dataset with pre-trained AlexNet as feature extractor.}\label{tb_vlcs_dataset}
\end{table}
Then, we examined our algorithm on the office-home benchmark, which is more challenging than the previous PACS and VLCS datasets. We follow the setting of~\citep{carlucci2019domain}, which is the most recent work who also evaluated on office-home dataset, to have a fair comparison. For this Office-home dataset, we also used reverse validation to set the learning rate as $2e-4$ for the whole model. For the remaining hyper-parameters, we keep the same with PACS and VLCS experiments. To avoid over-training, we also adopt the early stopping technique. All the experiments are programmed with \emph{PyTorch}~\citep{paszke2019pytorch}. 

% and tested on Nvidia RTX-2080Ti GPU with Intel Core-i9 9900K CPU.

\subsection{Experiments Results}

We first reported the empirical results on PACS and VLCS dataset using AlexNet as feature extractor in Table~\ref{tb_pacs_dataset} and Table~\ref{tb_vlcs_dataset}, respectively. For each generalization task, we train the model on all the source domains and test on the target domain and report the average of top 5 accuracy values. The empirical results refers to the average accuracy about training on source domains while testing on the target domain. 

From the empirical results, we can observe our method outperforms the baselines both on the PACS and VLCS dataset, indicating an improvement on benchmark performances. This showed the effectiveness of our method. Then, we report the empirical results on Office-home dataset in Table~\ref{WADG_results_office_home}. As stated before, Office-home is a more larger and challenging dataset contains more diverse features from $65$ different classes. To evaluate the performance on this dataset requires large amount of computational resources. Due to the limits, we follow the evaluation protocol of~\cite{carlucci2019domain} to report the empirical results. From those results, we could observe that our algorithm outperforms the previous Domain Generalization method, this also confirm the effectiveness of our proposed method.

%%%%%%%%%%%%%%%%%%5

\begin{table}[]
\centering

\resizebox{0.95\textwidth}{!}{\begin{tabular}{@{}l|llll|l@{}}
\toprule
         & Art   & Clipart & Product & Real-World & Avg.  \\ \midrule
Deep All & $52.15$ & $45.86$   & $70.86$   & $73.15$      & $60.51$ \\
D-SAM\citep{d2018domain}    & $\mathbf{58.03}$ & $44.37$   & $69.22$   & $71.45$      & $60.77$ \\
JiGen\citep{carlucci2019domain}    & $53.04$ & $\mathbf{47.51}$   & $71.47$   & $72.79$      & $61.20$ \\ \midrule
Ours     &   $55.34$    &  $44.82$       &  $\mathbf{72.03}$       &   $\mathbf{73.55}$         & $\mathbf{61.44}$      \\ \bottomrule
\end{tabular}}
\caption{Empirical Results (accuracy $\%$) on Office-home dataset with pre-trained ResNet-18 as feature extractor.}
\label{WADG_results_office_home}
\end{table}

\subsection{Further Analysis}
\label{further_analysis}
To further show the effectiveness of our algorithm especially on more deep models, follow~\cite{dou2019domain}, we also report the results of our algorithm by using ResNet-18 backbone on PACS dataset in Table~\ref{pacs_res18}. The ResNet-18 backbone, the output feature dim will be $512$. From the results, we could observe that our method could outperform the baselines on most generalization tasks and on average $+1.6\%$ accuracy improvement.

Then, we implement ablation studies on each component of our algorithm. We report the empirical results of ablation studies in Table~\ref{table_ablation_study}, where we test the ablation studies on both the AlexNet backbone and ResNet-18 backbone. We compare the ablations by, (1)~\emph{Deep All}: Train the model using feature extractor on source domain datasets with classification loss only, that is, neither optimal transport nor metric learning techniques is adopted.
(2)~\emph{No $\mathcal{L}_D$}: Train the model with classification loss and metric learning loss but without adversarial training component; {(3)~$\mathcal{L}_{MS}$ \emph{w.o.} $w^+$: omit the positive weighting scheme in $\mathcal{L}_{MS}$ (4) ~$\mathcal{L}_{MS}$ \emph{w.o.} $w^-$ : omit the positive weighting scheme in $\mathcal{L}_{MS}$.} (5)~\emph{No $\mathcal{L}_{MS}$}: Train the model with classification loss and adversarial loss but without metric learning component; (6)~\emph{WADG-All}: Train the model with full objective Eq.~\ref{Eq:full_objective}. 

\begin{table}[]
\centering
\caption{Empirical Results (accuracy $\%$) on PACS dataset with pre-trained ResNet-18 as feature extractor .}\label{pacs_res18}
\resizebox{1.0\textwidth}{!}{\begin{tabular}{l|llll|l}

\toprule
Method   &  Art   & Cartoon  & Sketch & Photo & Avg. \\
\hline
Deep All & $77.87$ & $75.89$ & $69.27$ & $95.19$ & $79.55$ \\ 
D-SAM\citep{d2018domain}    & $77.33$ & $72.43$ & $77.83$ & $95.30$ & $80.72$ \\
JiGen\citep{carlucci2019domain}     & $79.42$ & $75.25$ & $71.35$ & $96.03$ & $80.51$ \\
MASF\citep{dou2019domain}     & $80.29$ & $77.17$ & $71.69$ & $94.99$ & $81.04$ \\ 
MMLD\citep{dg_mmld}     & $81.28$ & $77.16$ & $72.29$ & $\mathbf{96.09}$ & $81.83$ \\ \bottomrule
Ours & $\mathbf{81.56}$ & $\mathbf{78.02}$ & $\mathbf{78.43}$ & $95.82$ & $\mathbf{83.45}$ \\
\bottomrule
\end{tabular}}
\end{table}

\begin{figure}
\centering
\begin{subfigure}{.40\textwidth}
  \centering
  % include first image
  \includegraphics[width=\textwidth]{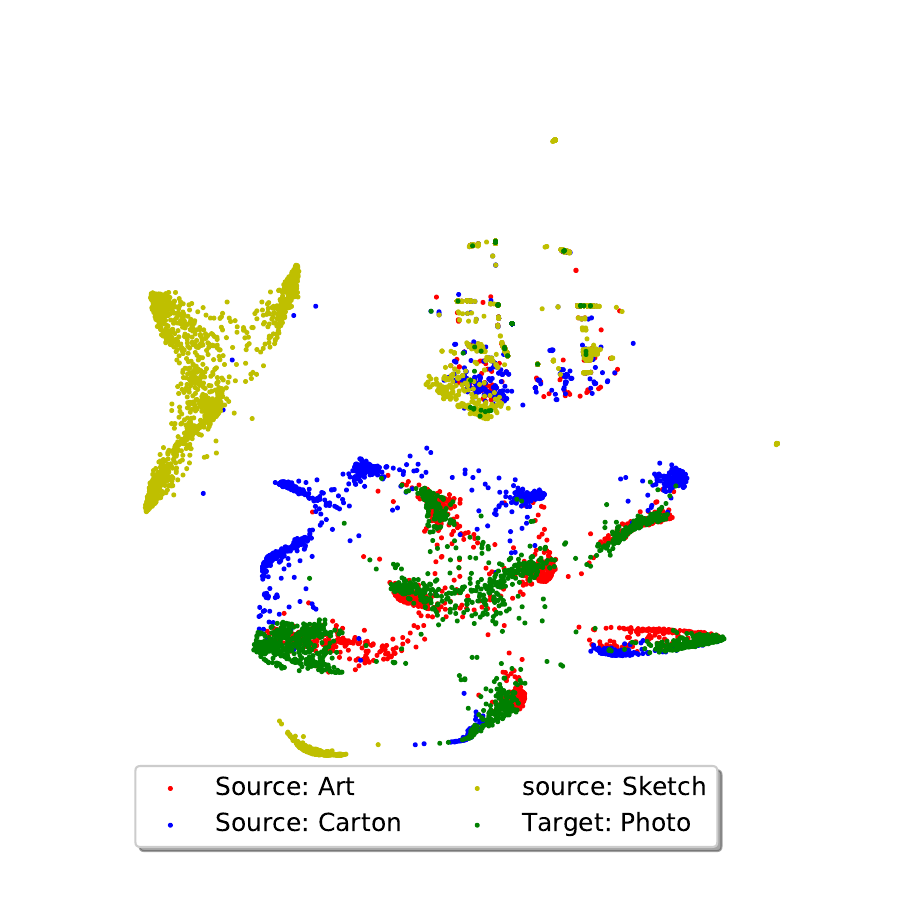}  
  \caption{Deep All}
  \label{fig:sub-first}
\end{subfigure}
\begin{subfigure}{.45\textwidth}
  \centering  % include second image
  \includegraphics[width=\textwidth]{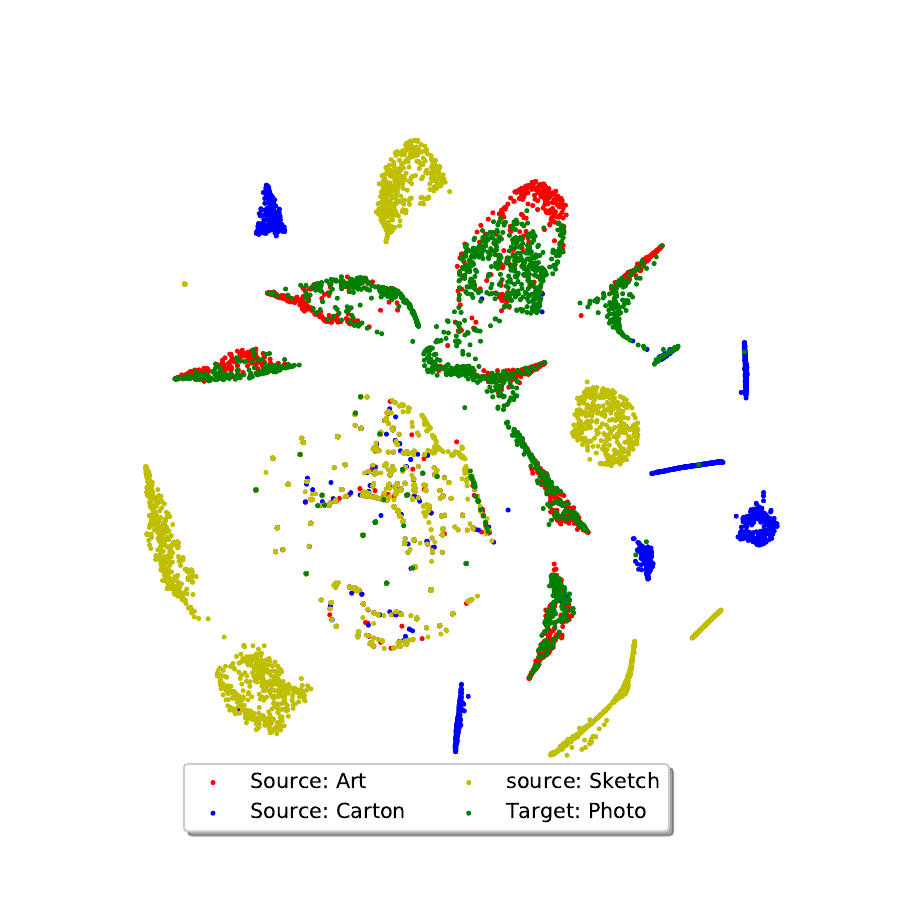}  
  \caption{No $\mathcal{L}_D$}
  \label{fig:sub-second}
\end{subfigure}\\
\begin{subfigure}{.45\textwidth}
  \centering
  % include third image
  \includegraphics[width=\textwidth]{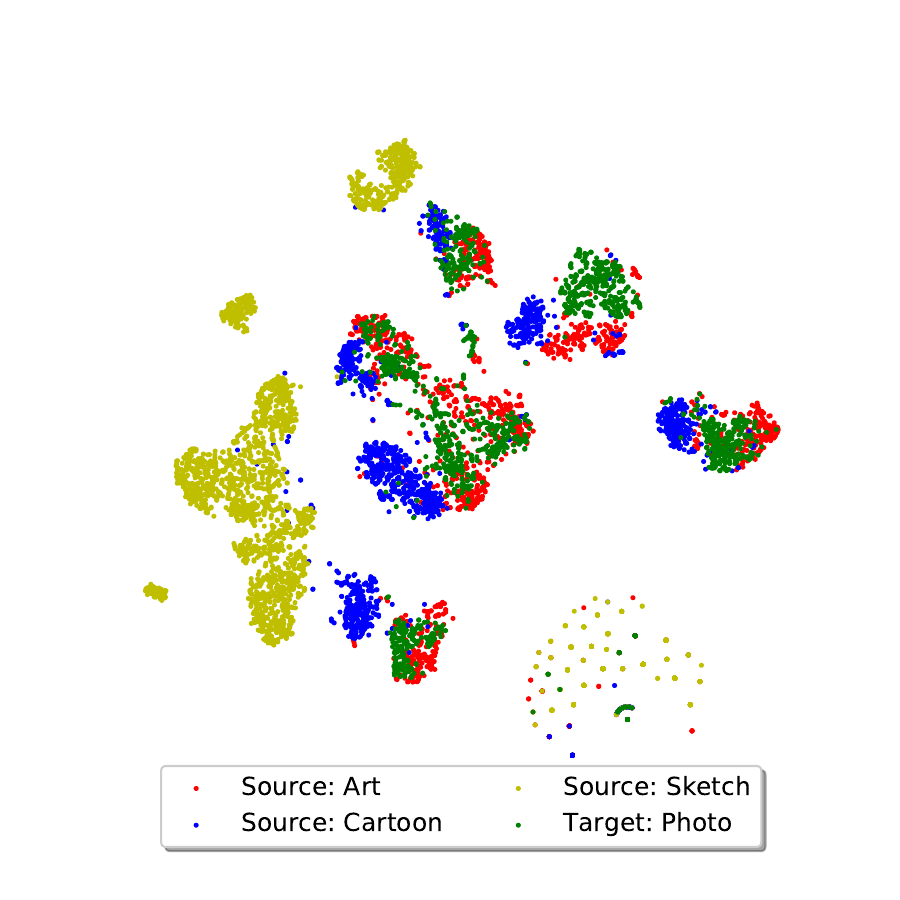}  
  \caption{No $\mathcal{L}_{MS}$}
  \label{fig:sub-third}
\end{subfigure}
\begin{subfigure}{.45\textwidth}
  \centering
  % include fourth image
  \includegraphics[width=\textwidth]{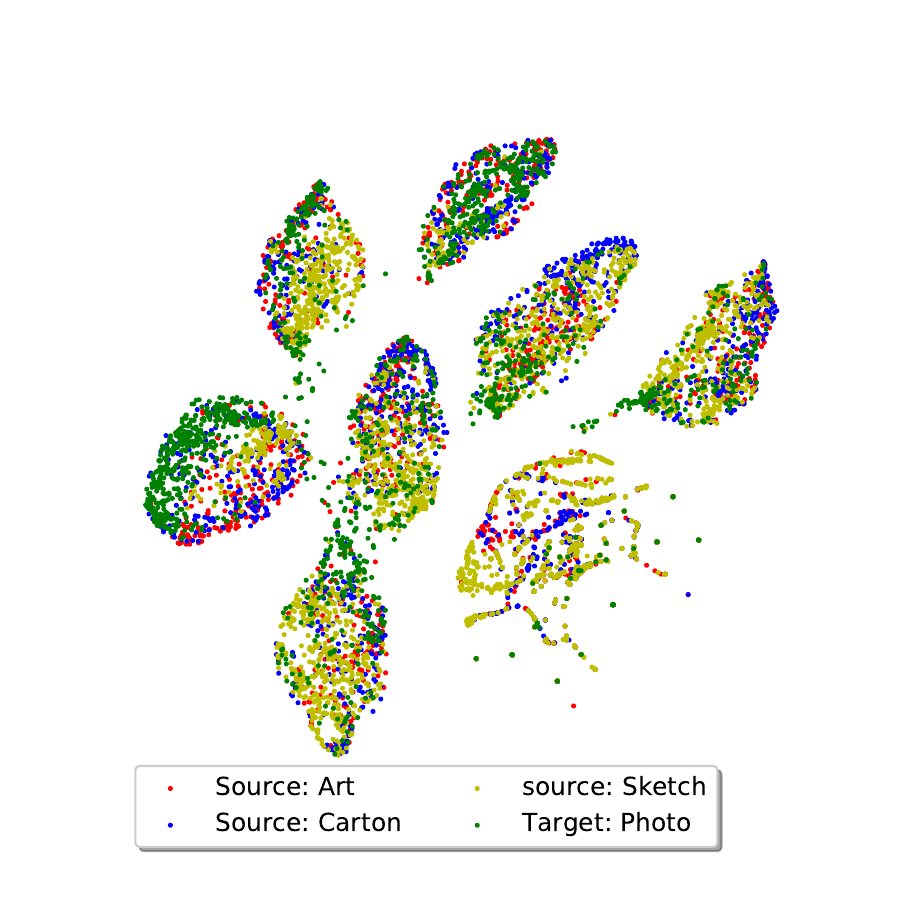}  
  \caption{WADG-All}
  \label{fig:sub-fourth}
\end{subfigure}
\caption{T-SNE visualization of ablation studies on PACS dataset for Target domain as \emph{Photo}. Detailed analysis is presented in section~\ref{further_analysis}. }
\label{fig:tsne}
\end{figure}

From the results, we could observe that one we omit the adversarial training, the accuracy would drop off rapidly ($\sim3.5\%$ with AlexNet backbone and $\sim5.8\%$ with ResNet-18 backbone). The contribution of the metric learning loss is relatively small compared with adversarial loss. {Comparing the ablations ~$\mathcal{L}_{MS}$ \emph{w.o.} $w^+$ and ~$\mathcal{L}_{MS}$ \emph{w.o.} $w^-$, we could observe almost similar accuracy. This indicates that the positive and negative weighting scheme of the metric learning objective may have equivalent contribution. }. Once we omit the metric learning loss, the performance will drop $\sim2.1\%$ and $\sim 2.5\%$ with AlexNet and ResNet-18 backbone, respectively.
%%%%%%%%%%5

%%%%%%%%%%%
\begin{figure}
\centering
\begin{subfigure}{.40\textwidth}
  \centering
  % include first image
  \includegraphics[width=\textwidth]{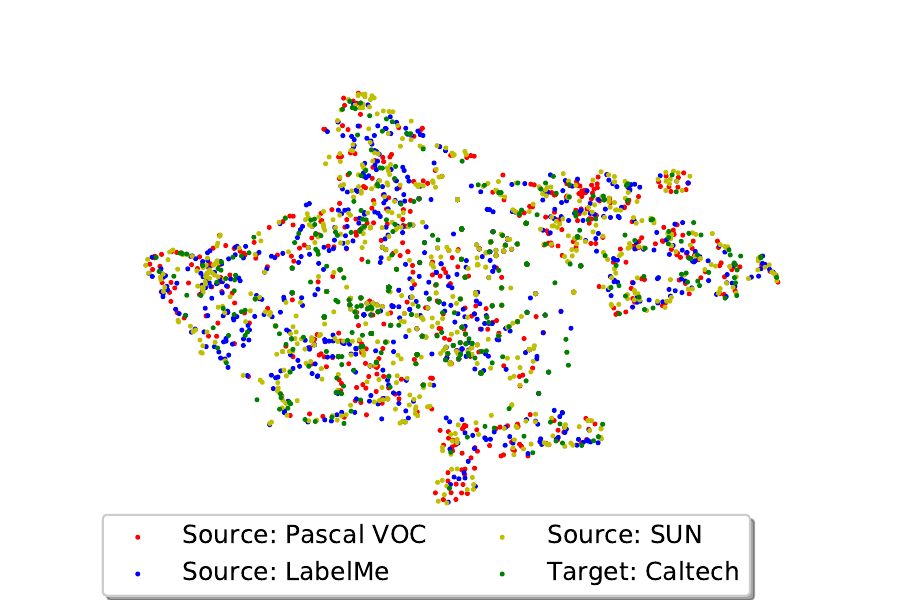}  
  \caption{Deep All}
  \label{fig:sub-first_VLCS}
\end{subfigure}
\begin{subfigure}{.45\textwidth}
  \centering  % include second image
  \includegraphics[width=\textwidth]{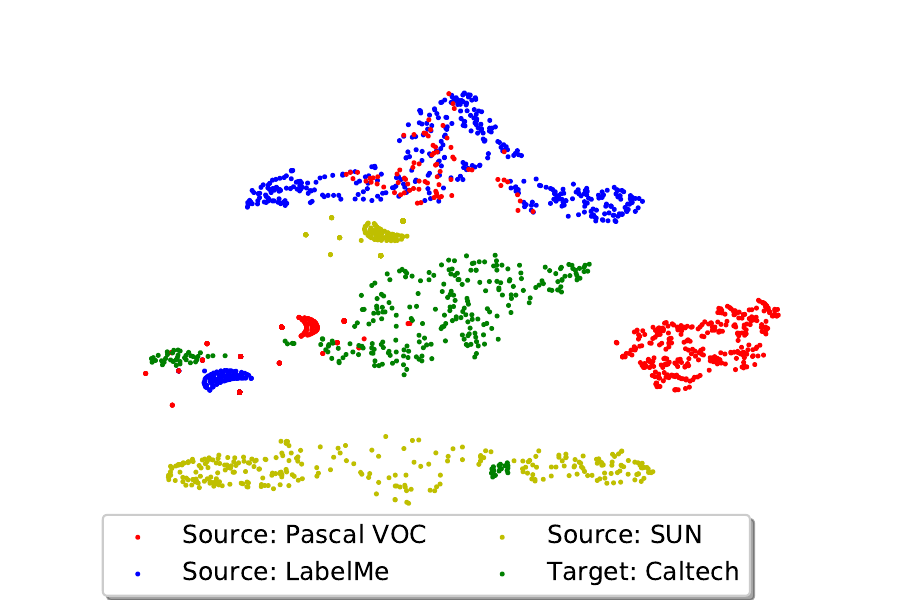}  
  \caption{No $\mathcal{L}_D$}
  \label{fig:sub-second_VLCS}
\end{subfigure}\\
\begin{subfigure}{.45\textwidth}
  \centering
  % include third image
  \includegraphics[width=\textwidth]{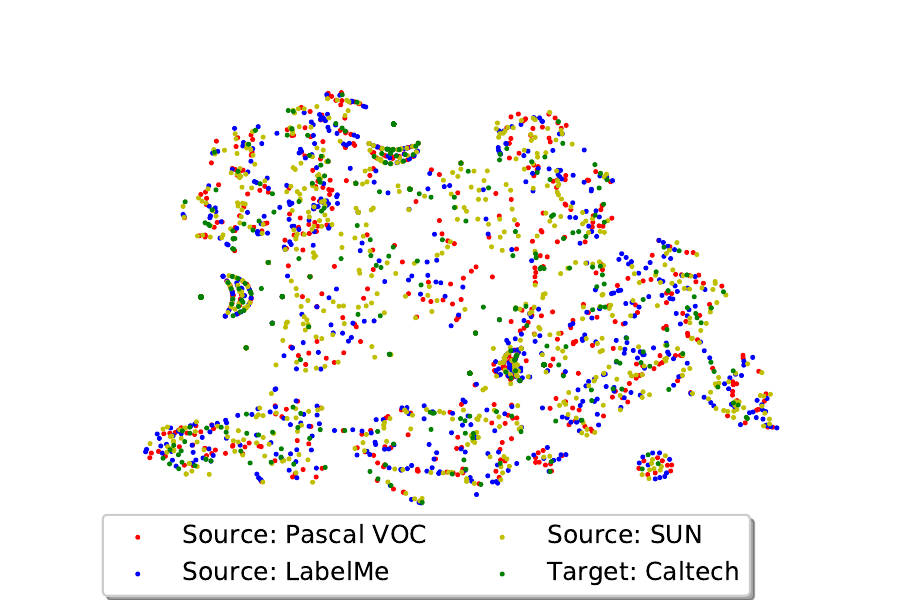}  
  \caption{No $\mathcal{L}_{MS}$}
  \label{fig:sub-third_VLCS}
\end{subfigure}
\begin{subfigure}{.45\textwidth}
  \centering
  % include fourth image
  \includegraphics[width=\textwidth]{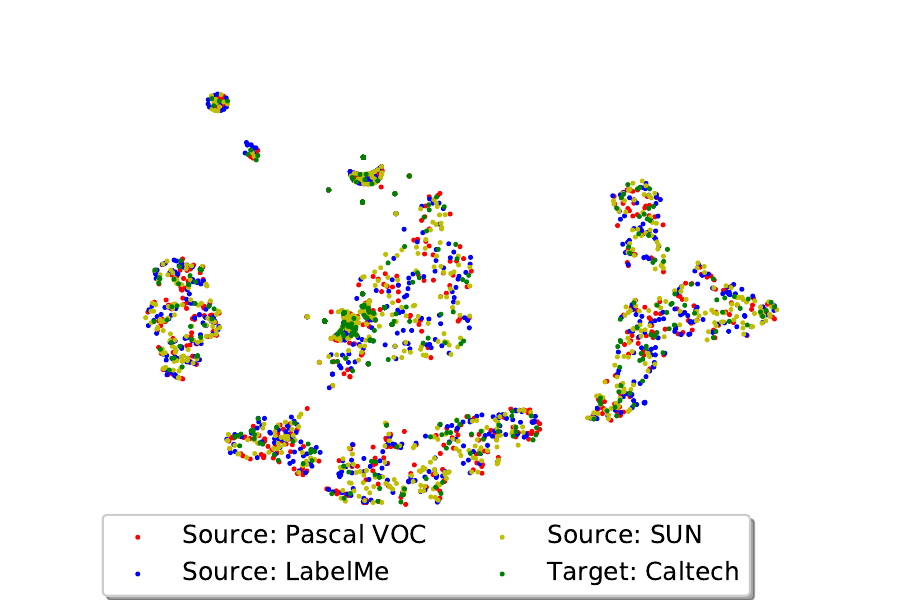}  
  \caption{WADG-All}
  \label{fig:sub-fourth_VLCS}
\end{subfigure}
\caption{T-SNE visualization of ablation studies on VLCS dataset for Target domain as \emph{Caltech}. Detailed analysis is presented in section~\ref{further_analysis}. }
\label{fig:tsne_VLCS}
\end{figure}

%%%%%%%5

\begin{table}[]

\centering
\caption{Ablation Studies on PACS dataset on all components of our proposed method using AlexNet and ResNet-18 backbone}
\label{table_ablation_study}
\resizebox{1.0\textwidth}{!}{\begin{tabular}{l|lllllllll|l}
\toprule
         & \multicolumn{5}{c|}{AlexNet}                                         & \multicolumn{5}{c}{ResNet-18}        \\
Ablation & Art     & Carton  & Sketch  & Photo   & \multicolumn{1}{l|}{Avg.}    & Art & Carton & Sketch & Photo & Avg. \\ \midrule
Deep All & $63.30$ & $63.13$ & $54.07$ & $87.70$ & \multicolumn{1}{l|}{$67.05$} & $77.87$ & $75.89$ & $69.27$ & $95.19$ & $79.55$ \\
No $\mathcal{L}_D$  & $65.80$ & $69.64$ & $63.91$ & $89.53$ & \multicolumn{1}{l|}{$72.22$}        & $74.62$    & $73.02$       & $68.67$       &  $94.86$     & $77.79$     \\
No $\mathcal{L}_{MS}$  & $66.78$ & $71.47$ & $68.12$ & $88.87$ & \multicolumn{1}{l|}{$73.65$}        &  $78.25$   &    $76.27$    & $73.42$       &    $95.68$   &  $80.91$    \\
{$\mathcal{L}_{MS}$ w.o. $w^{+}$} & { $66.31$}  & { $70.86$} & { $67.11$} & { $88.97$} & \multicolumn{1}{l|}{{ $73.31$}}        &  { $80.58$}   &   { $77.95$}    & { $75.13$}      &    { $95.63$}  & { $82.32$}   \\
{$\mathcal{L}_{MS}$ w.o. $w^{-}$} & { $66.41$}  & { $70.95$} & { $68.73$} & { $87.38$} & \multicolumn{1}{l|}{{ $73.37$}}        &  { $79.98$}   &   { $77.65$}    & { $77.89$}      &    { $95.21$}  & { $82.68$}   \\
WADG-All      &   $\mathbf{70.21}$      & $\mathbf{72.51}$        &    $\mathbf{70.32}$     &  $\mathbf{89.81}$       & \multicolumn{1}{l|}{$\mathbf{75.71}$} & $\mathbf{81.56}$    & $\mathbf{78.02}$       & $\mathbf{78.43}$       &  $\mathbf{95.82}$     &  $\mathbf{83.45}$    \\ \bottomrule
\end{tabular}}
\end{table}

Then, to better understand the contribution of each component of our algorithm, the T-SNE visualization of the ablation studies of each components on PACS {and VLCS} dataset are represented in Fig.~\ref{fig:tsne} for the generalization task of target domain \emph{Photo}. {and Fig.~\ref{fig:tsne_VLCS} for the generalization task of target domain \emph{Caltech}, respectively}. Since our goal is to not only align the feature distribution but also encourage a cohesion and separable boundary, in order to show the alignment and clustering performance, we report the T-SNE features of all the source domains and target domain to show the feature alignment and clustering across domains.\\
For PASC dataset, as we can see, the T-SNE features by \emph{Deep All} could neither project the instances from different domains to align with each other nor cluster the features into groups. The T-SNE features by \emph{No $\mathcal{L}_D$} showed the metric learning loss could to some extent to cluster the features, but without the adversarial training, the features could not be aligned well. The T-SNE features by \emph{No $\mathcal{L}_{MS}$} showed that the adversarial training could help to align the features from different domains but could not have a good clustering performance. The T-SNE features by \emph{WADG-All} showed that the full objective could help to not only align the features from different domains but also could cluster the features from different domains into several cluster groups, which confirms the effective of our algorithm. \\
{As for the VLCS dataset, we could observe similar performance on the T-SNE on the VLCS dataset while the features are somehow overlap with each other. This is due to the features in Caltech domain is somehow easy to learn and predict. As also analyzed in~\citep{Li2017dg}, a supervised model on Caltech domain could achieved $\sim 100\%$ accuracy, which also confirms that the features in Caltech domain is easy to learn indicating the features might be more likely overlapping with each other. As we can see from~Fig.\ref{fig:sub-fourth_VLCS}, the WADG method could help to separate the features with each other, which again confirms the effectiveness of our proposed method.}

\section{Conclusion}
In this paper, we proposed the Wasserstein Adversarial Domain Generalization algorithm for not only aligning the source domain features and transferring to an unseen target domain but also leveraging the label information across domains. We first adopt optimal transport with Wasserstein distance for aligning the marginal distribution and then adopt the metric learning method to encourage a domain-independent distinguishable feature space for a clear classification boundary. The experiments results showed our proposed algorithm could outperform most of the baseline methods on two standard benchmark datasets. Furthermore, the ablation studies and visualization of the T-SNE features also confirmed the effectiveness of our algorithm.   

\section*{Acknowledgement}
This work has been partially supported by Natural Sciences and Engineering Research Council of Canada (NSERC), The Fonds de recherche du Québec - Nature et technologies (FRQNT). Fan Zhou is supported by China Scholarship Council. Boyu Wang is supported by the Natural Sciences and Engineering Research Council of Canada (NSERC), Discovery Grants Program.
\\

\textbf{A full version of this preprint has been published as:}

Fan Zhou, Zhuqing Jiang, Changjian Shui, Boyu Wang, Brahim Chaib-draa,
Domain generalization via optimal transport with metric similarity learning, Neurocomputing,
Volume 456,
2021,
Pages 469-480,
https://doi.org/10.1016/j.neucom.2020.09.091.
(https://www.sciencedirect.com/science/article/pii/S0925231221002009)

\bibliography{mybibfile}
\end{document}